# Analysis of Sectoral Profitability of the Indian Stock Market Using an LSTM Regression Model

Jaydip Sen
Department of Data Science
Praxis Business School
Kolkata, INDIA
email: jaydip.sen@acm.org

Saikat Mondal
Department of Data Science
Praxis Business School
Kolkata, INDIA
email: saikatmondal15@gmail.com

Sidra Mehtab
Department of Data Science
Praxis Business School
Kolkata, INDIA
email: smehtab@acm.org

*Abstract*— Predictive model design for accurately predicting future stock prices has always been considered an interesting and challenging research problem. The task becomes complex due to the volatile and stochastic nature of the stock prices in the real world which is affected by numerous controllable and uncontrollable variables. This paper presents an optimized predictive model built on long-and-short-term memory (LSTM) architecture for automatically extracting past stock prices from the web over a specified time interval and predicting their future prices for a specified forecast horizon, and forecasts the future stock prices. The model is deployed for making buy and sell transactions based on its predicted results for 70 important stocks from seven different sectors listed in the National Stock Exchange (NSE) of India. The profitability of each sector is derived based on the total profit yielded by the stocks in that sector over a period from Jan 1, 2010 to Aug 26, 2021. The sectors are compared based on their profitability values. The prediction accuracy of the model is also evaluated for each sector. The results indicate that the model is highly accurate in predicting future stock prices.

## I. Introduction

Predictive model design for precise prediction of future stock prices has always been an interesting and challenging task due to the volatile and stochastic nature of the stock prices. Despite the proposition of the efficient market hypothesis regarding the impossibility of such prediction, there are propositions in the literature that have been able to demonstrate that future stock prices can be predicted by using advanced algorithms and predictive models. Time series decomposition and exponential smoothing are the two oldest methods for stock price prediction [1]. The use of machine learning and deep learning systems for stock price prediction have been the most popular approach in recent times [2-5]. Bollen et al. demonstrate how the emotions in the social web can significantly impact the volume of transactions in stock markets [3]. The proposal is based on computing the public sentiments and establishing a correlation between those moods and the DJIA index of the US stock market. Some works have also shown the capability of *convolutional neural networks* (CNN) for accurately forecasting stock prices [4].

Technical analysis of stocks for forecasting future prices is a very well-known approach. Several approaches to technical analysis have been proposed by researchers. These approaches are based on identifying well-known patterns in the time series of the stock price and forming appropriate investment strategies for deriving profit out of investments. A rich set of well-defined patterns and indicators have been proposed in the literature for this purpose.

The current work proposes a deep learning LSTM model for accurately predicting future stock prices. The model automatically extracts past prices of the stocks using a Python function that uses the ticker names of the stocks in NSE for an interval specified by a start date and an end date. Using the historical prices of 70 stocks from seven sectors the model is used to predict the future stock prices. Based on the prediction of the model, buy/sell decisions are taken for each stock and finally, the total profit earned for each stock is computed. Based on the aggregate profit of all the stocks for a given sector, the overall profitability of a sector is derived. A comparative analysis on the profitability of the seven sectors and the prediction accuracy of the LSTM model is evaluated.

The main contribution of the current work is threefold. First, an optimized deep learning model is proposed exploiting the power of LSTM architecture for predicting future stock prices for robust portfolio design. Second, the LSTM prediction is used as a guide in making buy/sell transactions on 70 stocks chosen from seven different sectors listed in NSE, India. The high precision of the model demonstrates its efficacy and effectiveness. Third, the returns of the stocks highlight the current profitability of investment and the volatilities of the seven sectors.

The paper is organized as follows. In Section II, some existing works on portfolio design and stock price prediction are discussed briefly. Section III presents a description of the methodology followed in this work. Section IV discusses the design of the proposed predictive model. Section V presents extensive results and their analysis. Section VI concludes the paper.

## II. Related Work

Researchers have worked extensively on various challenges in precisely predicting future stock prices and designing optimum portfolios for effectively trading off the associated risks and returns. Several approaches to time series decomposition, statistical and econometric modeling using methods like *autoregressive integrated moving average* (ARIMA), *autoregressive distributed lag* (ARDL), *vector autoregression* (VAR) are proposed in the literature for predicting future stock prices [6]-[21]. There has been extensive use of learning-based algorithms and architectures in an attempt to increase the prediction accuracy of stock prices and to enhance the robustness of stock portfolios [22]-[37]. Several propositions of hybrid models exist in the literature that combines the learning algorithms with sentiment information available in social media to arrive at more precise predictions [38]-[47]. Metaheuristics and several advanced heuristics approaches have also been proposed for solving the portfolio optimization problem [48]-[59]. For estimating and forecasting the future volatility of stock prices in the Indian stock market, the use of *generalized autoregressive conditional heteroscedasticity* (GARCH) is proposed in a recent work [60].

The current work proposes the use of a deep-learning regression model based on the LSTM architecture for accurately predicting future stock prices. Using the results predicted by the model, buy or sell decisions are made and the profit resulting from such buy/sell transactions are aggregated for a number of stocks which are selected from seven critical sectors listed in the NSE, India. The gross ratio of profit earned from the stock over a specified period to the mean price of the stock over the same period is taken as a metric for measuring the profitability of the stock. The average of this ratio for several stocks belonging to the same sector is considered as a measure of the profitability of the overall sector. Several sectors are analyzed and their profitability measures are computed. The analysis of the results provides a potential investor in the Indian stock market with valuable insights about the profitability of different sectors from the point of view of investment in stocks. Further, the work also demonstrates the effectiveness and efficacy of the predictive model.

III. METHODOLOGY

The methodology followed in this work involves five major steps. The steps are as follows: (1) *Data acquisition and loading,* (2) *Designing the model* (3) *Processing the output of the model,* (4) *Plotting the output for visualization,* and (5) *Predicting the future stock prices*. In what follows, the methods are discussed briefly.

**(1) Data acquisition and loading:** This step carries out extraction of the historical stock data from the Yahoo Finance website, performs all necessary preprocessing and transformation of the raw data. The stock prices are extracted based on their ticker names, the start date, and the end date for the data extraction. Python libraries are used for building the functions for performing this task. For all stocks, the start date is chosen as Jan 1, 2010, while the end date is Aug 26, 2021.

**(2) Designing the model:** In this step, the LSTM-based predictive model is designed. In designing the model, a function is used that uses the following parameters: (i) the length of the historical stock price data used as the input to the model, (ii) the number of variables (i.e., features) used in the input, (iii) the number of nodes in the LSTM layer, (iv) the number layers in the model, (v) the percentage of nodes used in dropout for regularizing the model, (vi) the type of loss function used in training and validation of the model, (vii) the optimizer used, (viii) the batch size used in training the model, and (ix) the number of epochs over which the model is trained. All these parameters are tunable. However, the following values have been used in the training and validation phase of the model.

The length of the historical stock price data used as n input is 50 implying past 50 days stock price records are used in making the next prediction. The number of input features used is five. These features are: open, high, low, close, and volume. The variable close is used as the variable to be predicted, while the others are used as the explanatory variables, also called the predictor variables. The default number of LSTM nodes used in the model is 256. The default value of the number of LSTM layers used is two. For achieving a sufficient degree of regularization of the model, the percentage of dropout is set to a default value of 30. The loss function is chosen to be the Huber loss due to its adaptive ability to handle complex data. The *Adam* optimizer is used as the default optimizer for optimizing the gradient descent algorithm used in the learning. A batch size of 64 and 100 epochs are used by default for training the LSTM model.

**(3) Processing the output of the model:** This step involves the execution of a Python function that works on two arguments, (i) the designed model in the previous step, and (ii) the transformed and pre-processed data. Based on the actual prices of the stock and the predicted prices returned by the model, the function performs some core tasks. These tasks are described briefly in the following.

In situations where the price predicted by the model for the next day is greater the today's price of the stock, the model will advise the investor to buy the stock today. The buy profit, however, is computed based on the actual price of the next day and not on the price that the model predicted on the previous day. The buy profits for all the days on which the investors sold the stock are summed up to arrive at the *total buy profit* for the stock.

However, when the model finds that the predicted price for the next is smaller than the price of today, a sell strategy is proposed for the current day, and the *selling profit* is computed as the difference between today's actual price of the stock and the actual price of the stock next day. The buy profits for all days on which the stock was bought are summed up to compute the *total sell profit* for the stock.

The *total profit* is computed by summing up the *total buy profit* with the *total sell profit*.

To evaluate the accuracy of the prediction of the model, this function also computes two metrics, the mean absolute error, and the accuracy score. While the former depicts the mean of the absolute values of the difference between the actual and predicted prices, the latter represents the percentage of cases in which the LSTM model could successfully predict the direction of movement (i.e., up or down) of the next day's price with respect to the price of the current day.

In the current work, there is an assumption of mandatory trading every day by a fictitious investor. When the forecasted price for the next day is higher than the current day's price, the investor compulsorily buys the stock, On the other hand, if the forecasted price is smaller or equal to the current day's price, then the investor goes for a mandatory selling of the stock. In the analysis, the buy/sell profits are computed on the assumption of buy/sell of a unit equity share. However, the rate of returns for the stocks and the sectors are not dependent on the number of equity shares transacted as those are percentage figures.

**(4) Plotting the output for visualization:** Several visualizations are made in this step. The plots which are particularly of importance are the training and validation loss plots of the model with respect to different epochs, and the plot of the actual and predicted stock prices by the LSTM.

**(5)** *Predicting the future stock prices:* This step involves a function to compute the future values of the stock prices based on the execution of the LSTM model. The forecast horizon is a tunable parameter of the function. However, in the current work, a forecast horizon of one day is used and the LSTM model predicted the price of the stock for the next day.

## IV. THE MODEL ARCHITECTURE

As explained in Section III, the stock prices are predicted with a forecast horizon of one day, using an LSTM model. This section presents the details of the architecture and the choice of various parameters in the model design. LSTM is an extended and advanced, *recurrent neural network* (RNN) with a high capability of interpreting and predicting future values of sequential data like time series of stock prices or text [61]. LSTM networks maintain their state information in some specially designed gates.

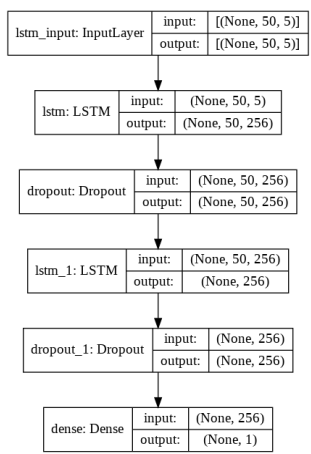

Fig. 1. The architecture of the LSTM model

For predicting the stock prices for the next day, an LSTM model is designed and fine-tuned. The design of the model is exhibited in Fig. 1. The model uses daily *close* prices of the stock of the past 50 days as the input. The input data of 50 days with a single feature (i.e., *close* values) is represented by the data shape of (50, 1). The input layer forwards the data to the first LSTM. The LSTM layer is composed of 256 nodes. The output from the LSTM layer has a shape of (50, 256). Thus, from every record in the input, the LSTM nodes extract 256 features. A dropout layer is used after the first LSTM layer that randomly switches off the output of thirty percent of the nodes to avoid model overfitting. Another LSTM layer with the same architecture as the previous one receives the output from the first and applies a dropout rate of thirty percent. A dense layer with 256 nodes receives the output from the second LSTM. The dense layer's output yields the predicted *close* price. The forecast horizon may be adjusted to different values by changing a tunable parameter. A forecast horizon of one day is used so that a prediction is made for the following day. The model is trained with a batch size of 64 and 100 epochs. With the exception of the final output layer, the ReLU is used as the activation function for all layers. The sigmoid function is used for activation in the final output layer. The loss and the accuracy are measured using the *Huber loss function* and the *mean absolute error* function, respectively. The grid search method is used to find out the optimum values of the hyperparameters of the model. The *Huber loss* function is used due to its superior ability to combine the features of MSE and MAE [61].

## V. EXPERIMENTAL RESULTS

Seven important sectors are selected from the sectors of stocks listed in the NSE, India. These sectors are: (i) energy, (ii) financial services, (iii) infrastructure, (iv) media, (v) pharmaceutical, (vi) private banks, and (vii) public sector banks. Based on the report of NSE released on July 30, 2021, the ten most significant stocks of these sectors are identified [63]. These stocks contribute most significantly in the computation of the sectoral index to which they belong. The LSTM model is deployed for computing the total profit (the sum of the buy and the sell profits) for each stock. The average of the ratio of the total profit to the mean price of each stock for the period of study (i.e., Jan 1, 2010 – Aug 26, 2021) is taken as the profitability metric for the sector. While the models are built on Python libraries, the execution was carried out on the Google Colab GPU [62]. The execution of one epoch required 3 seconds, on average.

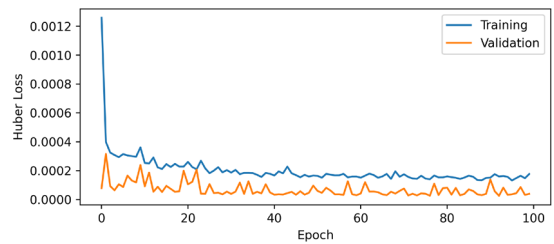

Fig. 2. The loss convergence of the LSTM model for Reliance Industries stock for different values of epoch

TABLE I. RESULTS OF THE ENERGY SECTOR

| Stock | Profit in Buying | Profit in Selling | Total Profit | Mean Price | Total Profit / Mean Price |
|---|---|---|---|---|---|
| RIL | 322650 | 321096 | 643746 | 439 | 1466 |
| PGC | 12622 | 12374 | 24996 | 90 | 278 |
| NTP | 8648 | 8570 | 17218 | 98 | 176 |
| ONG | 43508 | 43241 | 86748 | 84 | 1033 |
| BPC | 85710 | 87349 | 173059 | 116 | 1492 |
| AGE | 34516 | 34791 | 69307 | 462 | 150 |
| IOC | 27977 | 27614 | 55591 | 47 | 1183 |
| GAI | 33371 | 33119 | 33120 | 63 | 526 |
| TPC | 25455 | 25051 | 50506 | 48 | 1052 |
| HPC | 49586 | 50066 | 99652 | 74 | 1347 |
| Avg. profit/mean price of the sector: 870 | | | | | |

*Energy sector:* The ten most significant stocks in the energy sector listed in NSE are their weights in computing the sector index areas follows. Reliance Industries (RIL): 33.75, Power Grid Corporation of India (PGC): 11.90, NTPC (NTP): 11.42, Oil & Natural Gas Corporation (ONG): 8.55, Bharat Petroleum Corporation (BPC): 8.45, Adani Green Energy (AGE): 7.86, Indian Oil Corporation (IOC): 5.33, GAIL (GAI): 5.04, Tata Power Corporation (TPC): 4.30, and Hindustan Petroleum Corporation (HPC): 3.39. [63]. Table I presents the results for the stock of the energy sector. For each stock, the total profit in selling and buying is computed from Jan 1, 2005, till Aug 26, 2021. The total profit is divided by the mean price of the stock so that the profit earned from the stock is normalized over its price. The average of the ratios of the total profit to the mean price of all the ten stocks in the sector is finally computed to arrive at the overall profitability of the sector. The evaluation of loss in training and validation of the LSTM model over 100 epochs for the leading stock of the sector, Reliance Industries, is plotted in Fig. 2. The actual and predicted prices of the Reliance Industries stock by the LSTM model for the period Jan 1 to Aug 26, 2021, are plotted in Fig 3.

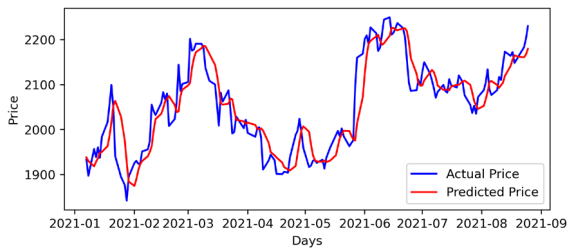

Fig. 3. The actual vs. the LSTM-predicted price of Reliance Industries stock from Jan 1, 2021, to Aug 26, 2021

*Financial Services sector:* As per the NSE report released on July 30, 2021, the ten most critical stocks in this sector and their respective weights in computing the sectoral index are as follows. HDFC Bank (HDF): 23.20, ICICI Bank (ICB): 17.61, Housing Development Finance Corporation (HDF): 16.40, Kotak Mahindra Bank (KMB): 9.05, Axis Bank (AXB): 6.80, State Bank of India (SBI): 6.18, Bajaj Finance (BJF): 6.16, Bajaj Finserv (BFS): 3.21, HDFC Life Insurance (HLI): 2.05, and SBI Life Insurance Company (SLI): 1.80 [63]. The results of this sector are presented in Table II. The training and validation loss for the LSTM model for HDFC Bank, the leading stock of this sector, are plotted in Fig. 4. The actual and predicted prices of the HDFC Bank stock by the LSTM model for the period Jan 1, 2021, and Aug 26, 2021, are depicted in Fig. 5.

TABLE II. RESULTS OF THE FINANCIAL SERVICES SECTOR

| Stock | Profit in Buying | Profit in Selling | Total Profit | Mean Price | Total Profit / Mean Price |
|---|---|---|---|---|---|
| HDB | 254876 | 255689 | 510565 | 308 | 1658 |
| ICB | 65854 | 65638 | 131492 | 196 | 671 |
| HDF | 351029 | 344886 | 695915 | 804 | 866 |
| KMB | 266915 | 266947 | 533862 | 486 | 1098 |
| AXB | 153804 | 153861 | 307665 | 250 | 1231 |
| SBI | 83296 | 82541 | 165837 | 141 | 1176 |
| BJF | 596418 | 592903 | 1189321 | 831 | 1431 |
| BFS | 1249371 | 1245330 | 2494701 | 2515 | 992 |
| HLI | 9873 | 9666 | 19539 | 532 | 37 |
| SLI | 15670 | 15247 | 30917 | 768 | 40 |
| Avg. profit/mean price of the sector:  920 | | | | | |

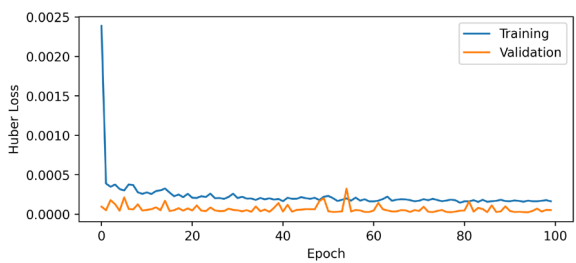

Fig. 4. The loss convergence of the LSTM model for HDFC Bank stock for different values of epoch

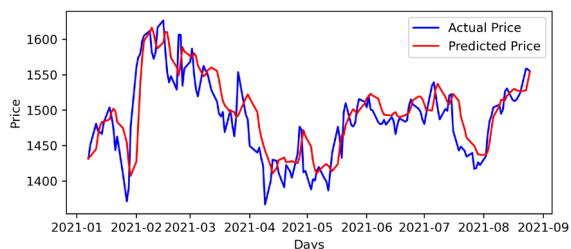

Fig. 5. The actual vs. the LSTM-predicted price of HDFC Bank stock from Jan 1, 2021, to Aug 26, 2021.

*Infrastructure sector:* The ten most significant stocks of this sector and their weights in the computation of the sectoral index are as follows. Reliance Industries (RIL): 19.10, Larsen & Toubro (LNT): 13.71, Bharti Airtel (BAL): 9.40, Ultra Tech Cement (UTC): 6.23, Grasim Industries (GSI): 4.20, Power Grid Corporation of India (PGC): 4.15, NTPC (NTP): 3.98, Adani Ports and Special Economic Zone (APZ): 3.51, Oil & Natural Gas Corporation (ONG): 2.98, and Bharat Petroleum Corporation (BPC): 2.95 [63]. The results of the *infrastructure* sector are presented in Table III. The training and validation loss for the LSTM model for Larsen & Toubro, the second most significant stock of this sector, are plotted in Fig. 6. It may be noted the plot for the first stock in this sector, Reliance Industries, is already presented in Fig. 7. The actual and predicted prices of the Larsen & Toubro stock by the LSTM model for the period Jan 1, 2021, and Aug 26, 2021, are depicted in Fig. 5.

TABLE III. RESULTS OF THE INFRASTRUCTURE SECTOR

| Stock | Profit in Buying | Profit in Selling | Total Profit | Mean Price | Total Profit / Mean Price |
|---|---|---|---|---|---|
| RIL | 322650 | 321096 | 643746 | 439 | 1466 |
| LNT | 223812 | 221486 | 445298 | 653 | 682 |
| BAL | 65615 | 64933 | 130548 | 283 | 461 |
| UTC | 775842 | 784734 | 1560576 | 2013 | 775 |
| GSI | 157608 | 158487 | 316095 | 475 | 665 |
| PGC | 12622 | 12374 | 24996 | 90 | 278 |
| NTP | 8648 | 8570 | 17218 | 98 | 176 |
| APZ | 52448 | 51735 | 104183 | 246 | 424 |
| ONG | 43508 | 43241 | 86748 | 84 | 1033 |
| BPC | 85710 | 87349 | 173059 | 116 | 1492 |
| Avg. profit/mean price of the sector:  745 | | | | | |

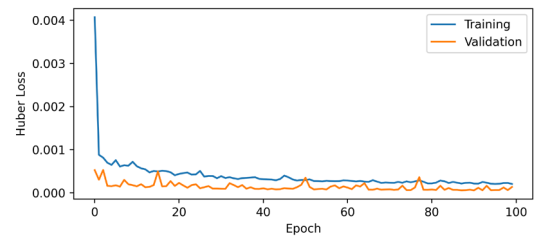

Fig. 6. The loss convergence of the LSTM model for Larsen & Toubro stock for different values of epoch

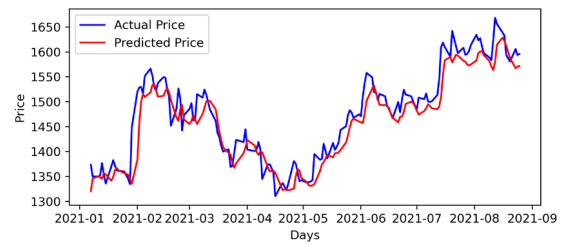

Fig. 7. The actual vs. the LSTM-predicted price of Larsen & Toubro stock from Jan 1, 2021, to Aug 26, 2021.

*Media sector:* The ten most significant stocks in the media sector and their weights used in the computation of the sectoral index, as per the report released by NSE on July 30, 2021, are as follows. Zee Entertainment Enterprises (ZEE): 28.28, PVR (PVR): 17.67, Sun TV Network (STN): 16.47, TV18 Broadcast (TVB): 10.38, Inox Leisure (INL): 8.49, Dish TV India (DTI): 6.92, Network 18 Media & Investments (NMI): 4.87, TV Today Network (TTN): 2.84, Jagran Prakashan (JPR): 2.33, and D. B. Corp (DBC): 1.76 [63]. Table IV presents the results of the *media* sector. The training and validation loss for the LSTM model for Zee Entertainment Enterprises, the leading stock of this sector, are plotted in Fig. 8. The actual and predicted prices of the

Zee Entertainment stock by the LSTM model for the period Jan 1, 2021, and Aug 26, 2021, are depicted in Fig. 9.

TABLE IV. RESULTS OF THE MEDIA SECTOR

| Stock | Profit in Buying | Profit in Selling | Total Profit | Mean Price | Total Profit / Mean Price |
|---|---|---|---|---|---|
| ZEE | 84692 | 83727 | 168419 | 205 | 822 |
| PVR | 232693 | 234510 | 467203 | 664 | 704 |
| STN | 65575 | 66486 | 132061 | 363 | 364 |
| TVB | 12270 | 12397 | 24667 | 52 | 474 |
| INL | 41658 | 41862 | 83520 | 159 | 525 |
| DTI | 10540 | 10389 | 20929 | 58 | 361 |
| NMI | 69066 | 69169 | 138235 | 135 | 1024 |
| TTN | 48825 | 48935 | 97760 | 154 | 635 |
| JPR | 16748 | 16687 | 33435 | 91 | 367 |
| DBC | 25947 | 25945 | 51892 | 226 | 230 |
| Avg. profit/mean price of the sector: 551 | | | | | |

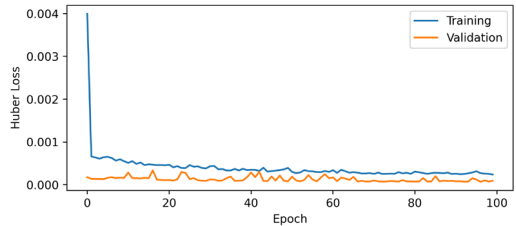

Fig. 8. The loss convergence of the LSTM model for the Zee Entertainment stock for different values of epoch

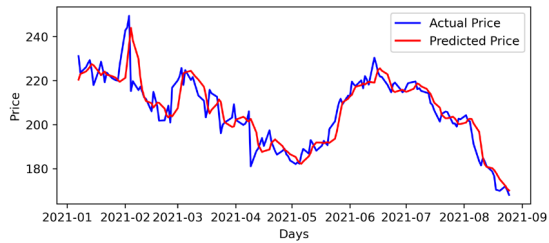

Fig. 9. The actual vs. the LSTM-predicted price of Zee Entertainment stock from Jan 1, 2021, to Aug 26, 2021.

*Pharmaceutical sector:* As per the report released by the NSE on July 30, 2021, the ten most significant stocks in this sector and their weights used in the computation of the sectoral index are as follows. Sun Pharmaceutical Industries (SPI): 22.87, Divi's Laboratories (DVL): 15.66, Dr. Reddy's Laboratories (DRL): 15.66, Cipla (CIP): 12.79, Lupin (LPN): 7.29, Aurobindo Pharma (ARP): 7.05, Biocon (BCN): 4.82, Alkem Laboratories (AKL): 4.18, Torrent Pharmaceuticals (TRP): 4.12, and Cadila Healthcare (CDH): 4.10 [63]. Table V presents the results of the *pharma* sector. The training and validation loss for the LSTM model for Sun Pharma, the leading stock of this sector, are plotted in Fig. 10. The actual and predicted prices of the Sun Pharma stock by the LSTM model for the period Jan 1, 2021, and Aug 26, 2021, are depicted in Fig. 11.

TABLE V. RESULTS OF THE PHARMACEUTICALS SECTOR

| Stock | Profit in Buying | Profit in Selling | Total Profit | Mean Price | Total Profit / Mean Price |
|---|---|---|---|---|---|
| SPI | 192050 | 190479 | 382529 | 247 | 1549 |
| DVL | 369872 | 368905 | 738777 | 742 | 996 |
| DRL | 913373 | 911228 | 1824601 | 1379 | 1323 |
| CPL | 165214 | 165690 | 330904 | 279 | 1186 |
| LPN | 282170 | 284315 | 566485 | 576 | 983 |
| APL | 186713 | 186215 | 372928 | 216 | 1727 |
| BCN | 47743 | 47644 | 95387 | 115 | 829 |
| AKL | 75534 | 77084 | 152618 | 2102 | 73 |
| TRP | 381069 | 382126 | 763195 | 689 | 1108 |
| CDH | 79284 | 79836 | 159120 | 172 | 925 |
| Avg. profit/mean price of the sector: 1070 | | | | | |

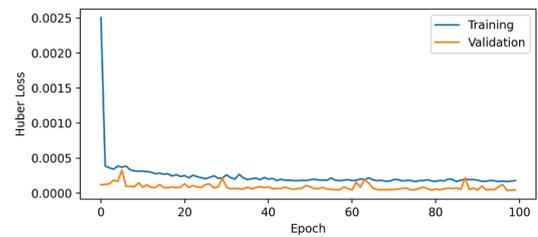

Fig. 10. The loss convergence of the LSTM model for the Sun Pharma stock for different values of epoch

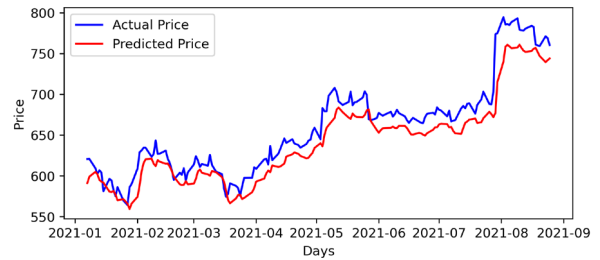

Fig. 11. The actual vs. the LSTM-predicted price of Sun Pharma stock from Jan 1, 2021, to Aug 26, 2021.

*Private Banks sector:* The ten most significant stocks of this sector and their weights used in deriving the sectoral index as per the NSE report released on July 30, 2021, are as follows. HDFC Bank (HDF): 25.47, ICICI Bank (ICB): 24.87, Kotak Mahindra Bank (KMB): 12.77, Axis Bank (AXB): 12.47, IndusInd Bank (ISB): 10.67, Bandhan Bank (BNB): 3.55, Federal Bank (FDB): 3.15, Yes Bank (YSB): 2.94, IDFC First Bank (IDF): 2.50, and RBL Bank (RBL): 1.62 [63]. Table VI presents the results of the *private banks* sector. The training and validation loss for the LSTM model for ICICI Bank, the second most critical stock of this sector, are plotted in Fig. 12. The actual and predicted prices of the ICICI Bank by the LSTM model for the period Jan 1, 2021, and Aug 26, 2021, are depicted in Fig. 13. The plots for the HDFC Bank stock, the leading stock of this sector are already depicted in Fig. 4 and Fig. 5 while discussing the performance of the *services* sector.

TABLE VI. RESULTS OF THE PRIVATE SECTOR BANKS

| Stock | Profit in Buying | Profit in Selling | Total Profit | Mean Price | Total Profit / Mean Price |
|---|---|---|---|---|---|
| HDB | 254876 | 255689 | 510565 | 308 | 1658 |
| ICB | 65854 | 65638 | 131492 | 196 | 671 |
| KMB | 266915 | 266947 | 533862 | 486 | 1098 |
| AXB | 153804 | 153861 | 307665 | 250 | 1231 |
| ISB | 287972 | 288941 | 576913 | 541 | 1066 |
| BNB | 11875 | 11685 | 23560 | 421 | 56 |
| FDB | 21838 | 21830 | 43668 | 30 | 1456 |
| YSB | 37030 | 36417 | 73447 | 98 | 749 |
| IDF | 1541 | 1535 | 3076 | 48 | 64 |
| RBL | 21946 | 21894 | 43840 | 407 | 108 |
| Avg. profit/mean price of the sector: 816 | | | | | |

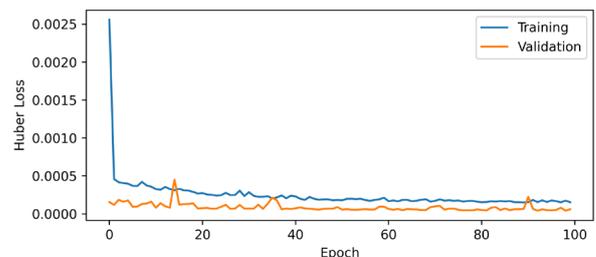

Fig. 12. The loss convergence of the LSTM model for ICICI Bank stock for different values of epoch

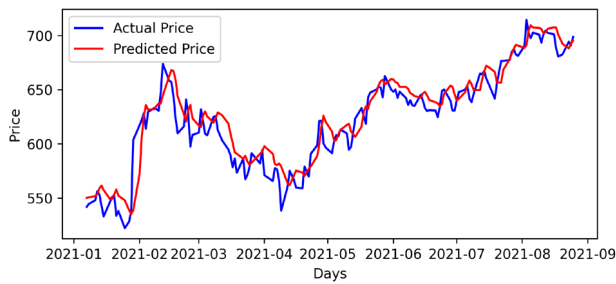

Fig. 13. The actual vs. the LSTM-predicted price of ICICI Bank stock from Jan 1, 2021, to Aug 26, 2021.

*Public Sector Banks:* The ten most significant stock and their weights used in computing the overall sectoral index as per the NSE report released July 30, 2021, are as follows. State Bank of India (SBI): 30.32, Bank of Baroda (BOB): 17.74, Punjab National Bank (PNB): 13.93, Canara Bank (CNB): 13.03, Indian Bank (INB): 5.77, Bank of India (BOI): 4.58, Union Bank of India (UBI): 4.20, Indian Overseas Bank (IOB): 3.08, Central Bank of India (CBI): 2.46, and Bank of Maharashtra (BMH): 1.68 [63]. Table VII presents the results of the public sector banks. The training and validation loss for the LSTM model for State Bank of India, the leading stock of this sector, are plotted in Fig. 14. The actual and predicted prices of the SBI stock by the LSTM model for the period Jan 1, 2021, and Aug 26, 2021, are depicted in Fig. 15.

TABLE VII. RESULTS OF THE PUBLIC SECTOR BANKS

| Stock | Profit in Buying | Profit in Selling | Total Profit | Mean Price | Total Profit / Mean Price |
|---|---|---|---|---|---|
| SBI | 83296 | 82541 | 165837 | 141 | 1176 |
| BOB | 30106 | 30217 | 60323 | 94 | 642 |
| PNB | 29002 | 28631 | 57633 | 102 | 565 |
| CNB | 57489 | 57585 | 115074 | 240 | 479 |
| INB | 28406 | 28501 | 56907 | 157 | 362 |
| BOI | 57519 | 56400 | 113919 | 168 | 678 |
| UBI | 35410 | 34854 | 70264 | 121 | 581 |
| IOB | 18456 | 18256 | 36712 | 53 | 693 |
| CBI | 12856 | 12852 | 25708 | 72 | 357 |
| BMH | 6316 | 6404 | 12720 | 31 | 410 |
| Avg. profit/mean price of the sector: 594 | | | | | |

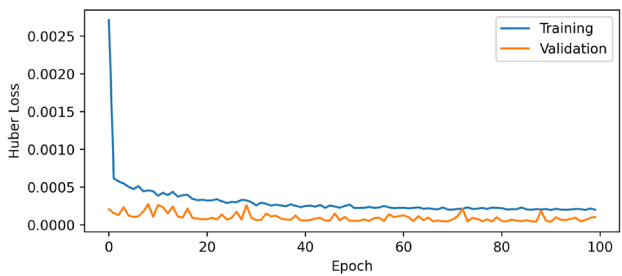

Fig. 14. The loss convergence of the LSTM model for the SBI stock for different values of epoch

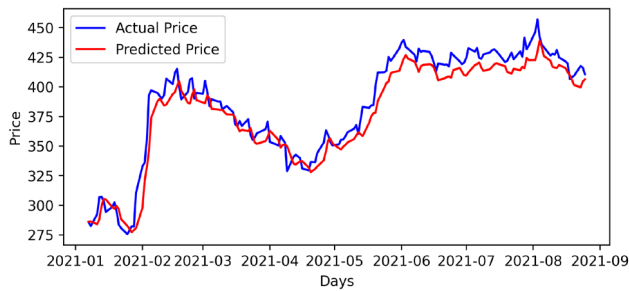

Fig. 15. The actual vs. the LSTM-predicted price of SBI stock from Jan 1, 2021, to Aug 26, 2021.

*Summary:* Since the metric that computes the average of the ratios of the total profit to the mean values of the prices of the constituent stocks of a sector represents the overall return on investment for the sector, it is used for comparing the profitability of the seven sectors that have been analyzed. It is observed that the pharmaceutical sector with a value of 1070 for the metric has the highest profitability, while the media sector is the least profitable with a value of 551 for the metric. Aurobindo Pharma stock yielded the highest profitability with a metric value of 1727, while HDFC Life Insurance was found to be the least profitable.

*Performance of the LSTM Model:* Finally, the performance of the LSTM is analyzed and presented. Three metrics are used for evaluating the performance of the LSTM model. The Huber loss (HL) is used to measure the loss exhibited by the model on the out-of-sample data (i.e., the test data), the mean absolute error (MAE) denotes the mean of the absolute values of the difference between the actual price and the predicted price of the stock by the LSTM model on the test data, and the accuracy score (AS) computes the percentage of cases for which LSTM predicted correctly the direction of movement of the stock price for the next day which was used by the investor to make buy/sell decisions. Tables VIII – XIV present the performance of the LSTM for the seven sectors studied in this work. The model yielded the lowest values for HL and MAE, for the pharma sector and the energy sector, respectively, while it produced the highest value of MAE for the pharma sector. Hence, the model has been the most accurate for these sectors on the three metrics. On the other hand, the private bank sector and the pharma sector exhibited the highest values for the metrics HL and MAE respectively, while AS was the least for the private sector banks.

TABLE VIII. LSTM PERFORMANCE ON ENERGY SECTOR

| Stock | Huber Loss | Mean Abs. Err. | Acc. Score |
|---|---|---|---|
| RIL | 0.000026 | 17.740238 | 0.982772 |
| PGC | 0.000073 | 32.441517 | 0.985097 |
| NTP | 0.000153 | 38.784095 | 0.975520 |
| ONG | 0.000044 | 5.590920 | 0.989788 |
| BPC | 0.000059 | 7.784823 | 0.974158 |
| AGE | 0.000270 | 43.544380 | 0.979591 |
| IOC | 0.000048 | 4.691075 | 0.993661 |
| GAI | 0.000056 | 5.435705 | 0.981848 |
| TPC | 0.000078 | 3.620325 | 0.992169 |
| HPC | 0.000062 | 8.699118 | 0.988254 |
| **Average** | **0.000087** | **16.833220** | **0.984286** |

TABLE IX. LSTM PERFORMANCE ON FINANCIAL SERVICES SECTOR

| Stock | Huber Loss | Mean Abs. Err. | Acc. Score |
|---|---|---|---|
| HDB | 0.000017 | 6.851545 | 0.992169 |
| ICB | 0.000043 | 18.879620 | 0.991462 |
| HDF | 0.000051 | 24.315115 | 0.992537 |
| KMB | 0.000038 | 10.726920 | 0.989889 |
| AXB | 0.000052 | 6.682780 | 0.983111 |
| SBI | 0.000064 | 12.863342 | 0.987480 |
| BJF | 0.000025 | 28.742176 | 0.953092 |
| BFS | 0.000021 | 144.067423 | 0.991416 |
| HLI | 0.000617 | 351.685975 | 0.988636 |
| SLI | 0.000526 | 522.597639 | 0.978022 |
| **Average** | **0.000145** | **112.741254** | **0.984781** |

TABLE X. LSTM PERFORMANCE ON INFRASTRUCTURE SECTOR

| Stock | Huber Loss | Mean Abs. Err. | Acc. Score |
|---|---|---|---|
| RIL | 0.000026 | 17.740238 | 0.982772 |
| LNT | 0.000060 | 39.3666145 | 0.988285 |
| BAL | 0.000077 | 14.211189 | 0.976546 |
| UTC | 0.000030 | 274.140146 | 0.989270 |
| GSI | 0.000048 | 43.921914 | 0.989339 |
| PGC | 0.000073 | 32.441517 | 0.985097 |
| NTP | 0.000153 | 38.784095 | 0.975520 |
| APZ | 0.000050 | 53.421672 | 0.974398 |
| ONG | 0.000044 | 5.590920 | 0.989788 |
| BPC | 0.000059 | 7.784823 | 0.974158 |
| **Average** | **0.000062** | **52.740313** | **0.982517** |

TABLE XI. LSTM PERFORMANCE ON MEDIA SECTOR

| Stock | Huber Loss | Mean Abs. Err. | Acc. Score |
|---|---|---|---|
| ZEE | 0.000068 | 20.988791 | 0.987207 |
| PVR | 0.000076 | 71.057895 | 0.977573 |
| STN | 0.000099 | 97.795559 | 0.981183 |
| TVB | 0.000097 | 13.785009 | 0.984507 |
| INL | 0.000081 | 24.253819 | 0.984043 |
| DTI | 0.000102 | 5.232211 | 0.988489 |
| NMI | 0.000150 | 18.672747 | 0.973233 |
| TTN | 0.000080 | 40.188656 | 0.987179 |
| JPR | 0.000085 | 22.151428 | 0.988032 |
| DBC | 0.000134 | 61.818736 | 0.989343 |
| **Average** | **0.000097** | **37.594485** | **0.984079** |

TABLE XII. LSTM PERFORMANCE ON PHARMACETICAL SECTOR

| Stock | Huber Loss | Mean Abs. Err. | Acc. Score |
|---|---|---|---|
| SPI | 0.000039 | 6.106475 | 0.994523 |
| DVL | 0.000017 | 23.620531 | 0.995570 |
| DRL | 0.000032 | 60.127631 | 0.990603 |
| CPL | 0.000039 | 9.493322 | 0.991366 |
| LPN | 0.000046 | 15.948761 | 0.996942 |
| APL | 0.000046 | 7.098335 | 0.987421 |
| BCN | 0.000044 | 15.722546 | 0.984634 |
| AKL | 0.000173 | 1171.005451 | 0.962963 |
| TRP | 0.000033 | 27.379740 | 0.988285 |
| CDH | 0.000033 | 27.379740 | 0.988285 |
| **Average** | **0.000050** | **136.388253** | **0.988059** |

TABLE XIII. LSTM PERFORMANCE ON PRIVATE BANKS SECTOR

| Stock | Huber Loss | Mean Abs. Err. | Acc. Score |
|---|---|---|---|
| HDB | 0.000017 | 6.851545 | 0.992169 |
| ICB | 0.000043 | 18.879620 | 0.991462 |
| KMB | 0.000038 | 10.726920 | 0.989889 |
| AXB | 0.000052 | 6.682780 | 0.983111 |
| ISB | 0.000049 | 21.770370 | 0.988248 |
| BNB | 0.000842 | 169.656113 | 0.949367 |
| FDB | 0.000048 | 1.172474 | 0.978774 |
| YSB | 0.000071 | 9.951633 | 0.987212 |
| IDF | 0.000281 | 19.174256 | 0.945652 |
| RBL | 0.000309 | 115.524336 | 0.983051 |
| **Average** | **0.000175** | **38.039005** | **0.978894** |

TABLE XIV. LSTM PERFORMANCE ON PUBLIC SECTOR BANKS

| Stock | Huber Loss | Mean Abs. Err. | Acc. Score |
|---|---|---|---|
| SBI | 0.000120 | 8.081225 | 0.984009 |
| BOB | 0.000120 | 8.081225 | 0.984009 |
| PNB | 0.000076 | 8.056016 | 0.990354 |
| CNB | 0.000088 | 35.559916 | 0.987952 |
| INB | 0.000099 | 45.528190 | 0.985735 |
| BOI | 0.000083 | 21.947678 | 0.988273 |
| UBI | 0.000081 | 12.192439 | 0.985885 |
| IOB | 0.000069 | 7.970885 | 0.990405 |
| CBI | 0.000079 | 11.657904 | 0.983776 |
| BMH | 0.000088 | 8.960729 | 0.988180 |
| **Average** | **0.000090** | **16.803621** | **0.986858** |

## VI. CONCLUSION

This paper has presented an LSTM model for predicting future stock prices. The model is optimized with suitably designed layers and regularized using the dropout regularization method. The historical stock prices for 70 stocks from seven different sectors listed in NSE, India are automatically extracted from the web from Jan 1, 2010, Aug 26, 2021. The model is used for predicting the future stock prices with a forecast horizon of 1 day, and based on the predicted output of the model buy/sell decisions are taken. The total profit earned from the buy/sell transactions for a stock is normalized by its mean price over the entire period to arrive at the profitability measure of the stock. The profitability figures of all stocks in a given sector are summed up to derive the overall profitability of the sector. It is observed that while the pharma sector is the most profitable one, the least profitable one is the media sector. The accuracy of the predictive model is measured using three metrics, Huber loss, mean absolute error (MAE), and the accuracy score. It is observed that the LSTM model is highly accurate in its prediction for all stocks analyzed in the study. As a future scope of work, some other sectors also will be studied for exploring their profitability and the accuracy of prediction of the LSTM model on the stocks of those sectors.